\documentclass[runningheads]{llncs}

\usepackage[T1]{fontenc}
\def\doi#1{\href{https://doi.org/\detokenize{#1}}{\url{https://doi.org/\detokenize{#1}}}}
\usepackage{graphicx}
\usepackage{listings}
\usepackage{amssymb} 
\usepackage{xcolor}
\usepackage{multirow}

\newcommand\rel{\mathcal{R}}

\begin{document}
\title{Learning video retrieval models with relevance-aware online mining} 
%
%
\author{Alex Falcon\inst{1,2}
\and
Giuseppe Serra\inst{2}
\and
Oswald Lanz\inst{3}
}
\authorrunning{Falcon et al.}
%
\institute{Fondazione Bruno Kessler, Povo TN 38123, Italy\\
\email{afalcon@fbk.eu}\and
University of Udine, Udine UD 33100, Italy\\
\email{giuseppe.serra@uniud.it}\and
Free University of Bozen-Bolzano, Bolzano BZ 39100, Italy\\
\email{lanz@inf.unibz.it}}
%
\maketitle              
\begin{abstract}
Due to the amount of videos and related captions uploaded every hour, deep learning-based solutions for cross-modal video retrieval are attracting more and more attention. A typical approach consists in learning a joint text-video embedding space, where the similarity of a video and its associated caption is maximized, whereas a lower similarity is enforced with all the other captions, called \emph{negatives}. This approach assumes that only the video and caption pairs in the dataset are valid, but different captions - \emph{positives} - may also describe its visual contents, hence some of them may be wrongly penalized. To address this shortcoming, we propose the \emph{Relevance-Aware Negatives and Positives} mining (RANP) which, based on the semantics of the negatives, improves their selection while also increasing the similarity of other valid positives. 
We explore the influence of these techniques on two video-text datasets: EPIC-Kitchens-100 and MSR-VTT. By using the proposed techniques, we achieve considerable improvements in terms of nDCG and mAP, leading to state-of-the-art results, e.g. +5.3\% nDCG and +3.0\% mAP on EPIC-Kitchens-100. We share code and pretrained models at \url{https://github.com/aranciokov/ranp}.
%
%
\keywords{Video retrieval \and cross-modal retrieval 
\and contrastive loss \and hard negative mining.}
\end{abstract}

\section{Introduction}
When performing a search by typing a textual query on a multimedia search engine, the user expects the retrieved contents to be semantically close to it. As one can expect, it is important for the first retrieved item to be `exactly' what the user was looking for. Yet, the following ones should be treated as importantly as the first one, given that multiple items are likely relevant to the user query. In a recent work, Wray et al. \cite{wray2021semantic} described this problem for the domain of video retrieval. Similar observations were also made in previous work in different domains, e.g. in image retrieval \cite{gordo2017beyond}. Focusing on text-video retrieval, most of the current methods learn a joint textual-visual embedding space (e.g. \cite{Chen_2020_CVPR,dong2021dual}). To do so, a video descriptor and a textual descriptor are computed independently for each of the pairs (video and its captions) in the dataset; then, the similarity of these descriptors is maximized. At inference time, given a textual query, the multimedia search engine would retrieve the related video as the first result. To achieve this goal, a typical choice consists in contrastive loss functions, e.g. \cite{gutmann2010noise,hadsell2006dimensionality}, which contrast the similarity of the paired video and text descriptors against those of different videos and captions called `negatives'.

Several techniques have been proposed to decide which negatives to use in order to drive the learning, e.g. `hard' or `semi-hard' \cite{hermans2017defense,schroff2015facenet}, how many of them, e.g. one \cite{schroff2015facenet}, two \cite{chen2017beyond}, or more \cite{sohn2016improved}. Yet, in all these cases the pool of negatives contains the captions which are not paired in the dataset: thus, the selection of the negatives is often unaware of the overlap between the semantic content of the caption and the contents of the video. As an example, a dataset may contain a video $v^\star$ and its caption $q^\star$ about a chef preparing and baking a cheesecake; a different caption $q_1$ about preparing a cheesecake without baking it; and $q_2$ describing how to change a light bulb. By using the aforementioned strategies, during training $q_1$ and $q_2$ may be selected as negative captions, and their similarity to $v^\star$ and $q^\star$ would be lowered. While this is fine for $q_2$, $q_1$ should be treated differently considering its similarity with $q^\star$. To address this shortcoming, in this paper we propose to improve the mining by making it aware of such an overlap. 
In particular, we focus on `online' mining (i.e. the negatives are picked from the batch) because it is widely used in recent works (e.g. \cite{Chen_2020_CVPR,dong2021dual}) and is less burdensome than `offline' mining. Moreover, to estimate the overlap we use a relevance function \cite{damen2020rescaling} defined on already available captions, avoiding the need for costly annotations. Thus, we name it relevance-aware online negative mining, or RAN. Differently from previous techniques which might select less false negatives after training for some epochs, RAN improves the selection from the start thanks to the use of semantics and not only the network state. 

Similarly, we extend this idea to select captions which present a considerable overlap with the video contents (`positive' captions), but are not paired to the video. In fact, video retrieval methods which perform online mining only consider the groundtruth pairs as valid positives, missing this opportunity. A few works using offline mining select them based on semantic class labels \cite{wray2021semantic,wray2019fine} and, in different domains, this is done by using additional data which are not available in video-text datasets, e.g. class identifiers for image retrieval \cite{xuan2020improved} and person re-identification \cite{hermans2017defense}. By merging this technique with RAN, we obtain RANP which carefully mines both negative and positive captions using semantics. 

The main contributions of this work can be summarized as follows:
\begin{itemize}
    \item to address the shortcoming of false negatives' selection during the online hard negative mining, we propose and formulate a relevance-aware variant that we call RAN which, differently from previous techniques, uses semantics to select better negatives from the start of the training;
    \item we introduce in the video retrieval field the relevance-aware online hard positive mining, which helps selecting hard positives thanks to a relevance-aware mechanism, and use it alongside RAN obtaining RANP;
    \item we validate the proposed techniques on two public video-and-language datasets, which are EPIC-Kitchens-100 and MSR-VTT, providing evidence of their usefulness while also achieving the new state-of-the-art on EPIC-Kitchens-100 with an improvement of +5.3\% nDCG and +3.0\% mAP.
\end{itemize}

\section{Related work}
\textbf{Video retrieval.} 
The approaches for cross-modal text-video retrieval usually learn a joint textual-visual embedding space \cite{Chen_2020_CVPR,croitoru2021teachtext,dong2021dual,wang2021t2vlad}. Given the multimodal nature of videos, several authors introduced novel techniques to learn a joint representation of all the available modalities \cite{gabeur2020multi,liu2019use,miech2018learning,mithun2018learning,wang2021t2vlad}. As an example, Mixture of Embedding Experts (or MoEE, by Miech et al. \cite{miech2018learning}) and T2Vlad (by Wang et al. \cite{wang2021t2vlad}) applied techniques based on NetVLAD \cite{arandjelovic2016netvlad}, whereas a multimodal Transformer was used in Gabeur et al. \cite{gabeur2020multi}. Liu et al \cite{liu2019use} proposed Collaborative Experts (CE), which extended previous works with a gating mechanism to modulate each feature based on the other pretrained experts. Recently, Croitoru et al. \cite{croitoru2021teachtext} shifted the attention to the textual counterpart, leveraging the availability of multiple language models. 
The structure of the input data was used in multiple works, e.g. by constructing embedding spaces based on the part-of-speech (Wray et al. \cite{wray2019fine}) or by employing semantic role labeling to learn global and local representations (Chen et al. \cite{Chen_2020_CVPR}).
All these works focus on instance-based video retrieval, where only video and caption pairs in the dataset are considered to evaluate the performance. Given that multiple descriptions can describe a video, Wray et al. \cite{wray2021semantic} proposed a `semantic' video retrieval, which considers multiple degrees of relevance when computing the evaluation metrics. 

\textbf{Contrastive loss and mining techniques.} To learn the cross-modal embedding spaces, contrastive losses \cite{gutmann2010noise,hadsell2006dimensionality,hermans2017defense} are often employed because they enforce a high similarity for the descriptors of (video, caption) pairs in the dataset. 
Hadsell et al. \cite{hadsell2006dimensionality} initially computed the loss on pair of samples, and the idea has been extended to triplets \cite{schroff2015facenet}, quadruplets \cite{chen2017beyond}, and `N+1'-tuples \cite{sohn2016improved}. Yet, the amount of possible tuples scales exponentially (e.g. cubically with triplets) and most of them contribute meaninglessly to the loss. Hence, mining techniques were proposed to extract less tuples, either from the dataset (`offline') or from the batch (`online'). Offline mining is often avoided because it recomputes the tuples throughout the training making it burdensome. Nonetheless, some works made use of it in several domains, e.g. deep metric learning \cite{harwood2017smart,suh2019stochastic} and video retrieval \cite{wray2021semantic,wray2019fine}. 
In `online' mining, the positive items are given by groundtruth associations, e.g. (video, caption) pairs in the dataset, whereas the rest of the batch forms the pool of negatives. The loss is often computed on all negatives (e.g. \cite{gabeur2020multi,miech2018learning}), but picking `hard' or `semi-hard' negatives (i.e. irrelevant but highly similar to the groundtruth) is often preferred, as in \cite{Chen_2020_CVPR,dong2021dual}. Nonetheless, recent research (e.g. Xuan et al. \cite{xuan2020hard,xuan2020improved}) presented the usefulness of easy examples. Mining techniques for positive items have also been proposed (e.g. in cross-modal \cite{hermans2017defense,xuan2020improved} and near-duplicate video retrieval \cite{jiang2019svd}) although they are based on the availability of groundtruth labels. In representation learning for images or videos, positive items were also artificially constructed through transformations \cite{chen2020simple,he2020momentum,pan2021videomoco,qian2021spatiotemporal}. 

We aim at introducing semantic knowledge in training by leveraging the relevance function. A similar idea is also used in \cite{wray2021semantic}, but we do so to improve online mining techniques, which are usually preferred. Furthermore, differently from previous video retrieval methods, we present a two-step approach to select online hard positives and show its effectiveness on two large scale datasets.

\section{Training a video retrieval model with contrastive loss and mining}
\begin{figure}[t]
    \centering
    \includegraphics[width=\linewidth]{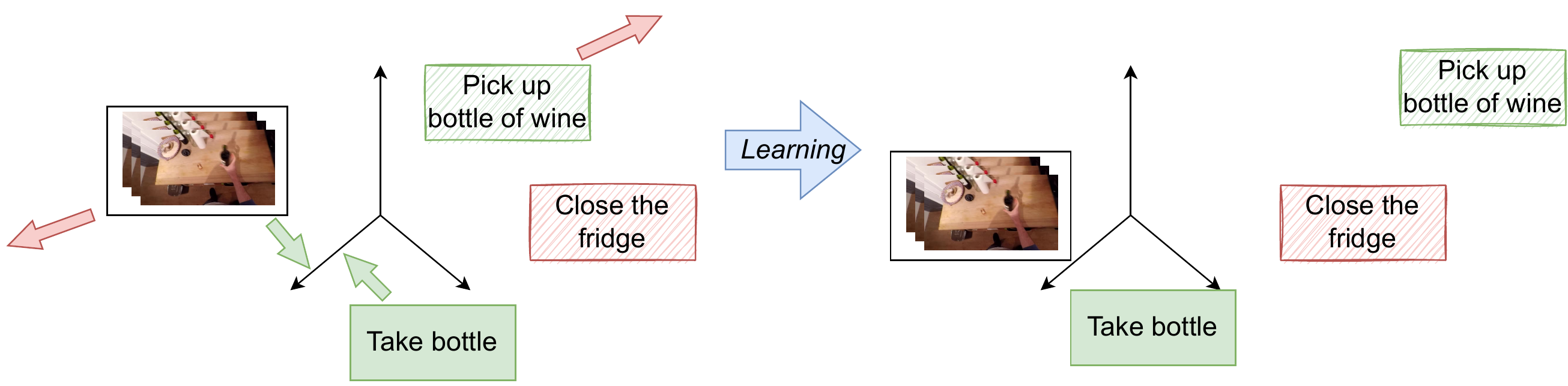}
    \caption{By adopting the typical approach (Eq.~\ref{eq:hardneg_standard}), the caption `pick up bottle of wine' is used as the hardest negative because it is not associated to the video in the dataset, and its descriptor is the most similar (closest) to the video. We visualize the change in similarity with the green (increase) and red (decrease) arrows. }
    \label{fig:hardneg_visual}
\end{figure}
Given a video $v^\star$ and a pool of candidate textual queries, the aim of video-to-text retrieval is to return a ranked list of candidates where on top we expect the caption $q^\star$ corresponding to $v^\star$ in the dataset. `Ranked list' implies that the output is sorted based on the similarity (computed with $s(\cdot, \cdot)$, e.g. cosine similarity) of $v^\star$ with all the candidate captions. In a common video retrieval setting, the evaluation metrics solely focus on the rank of the corresponding caption. But, multiple captions may equally describe the same video, so we focus on semantic video retrieval \cite{wray2021semantic}, where the evaluation is based on metrics which look at the whole ranked list. In the rest of the paper we focus on video-to-text, but text-to-video retrieval is obtained by swapping the role of $q^\star$ and $v^\star$. 

To train the video retrieval model, a contrastive loss which performs online mining of triplets \cite{schroff2015facenet} is widely used \cite{Chen_2020_CVPR,dong2021dual,dzabraev2021mdmmt,gabeur2020multi} and is based on this term:
\begin{equation}
    L_n = max(0, \Delta_n + s(v^\star, q-) - s(v^\star, q^\star))\label{eq:triplet_loss_term}
\end{equation}
where $\Delta_n$ is a fixed margin, 
and $q-$ is a query which does not describe the video $v^\star$ (i.e. $q-$ is a negative query for $v^\star$). By optimizing with respect to Eq.~\ref{eq:triplet_loss_term}, a margin $\Delta_n$ is enforced between the similarity of the groundtruth pair and the similarity of video and negative query, in order to satisfy the following constraint:
\begin{equation}
    s(v^\star, q-) + \Delta_n < s(v^\star, q^\star)\label{eq:triplet_loss_constraint}
\end{equation}
Note that Eq.~\ref{eq:triplet_loss_term} can be optimized on the whole mini-batch (e.g. in \cite{gabeur2020multi,miech2018learning}), but this leads to the inclusion of several easy negative captions (i.e. already satisfying Eq.~\ref{eq:triplet_loss_constraint}) which do not provide a meaningful contribution to the loss. Hence, to pick useful triplets, online hard negative mining is often preferred (e.g. \cite{Chen_2020_CVPR,dong2021dual}).

\subsection{Online hard negative mining\label{sec:online_hardneg_mining}}
Online hard negative mining consists in optimizing Eq.~\ref{eq:triplet_loss_term} only on hard negatives, which are the captions $q$ violating Eq.~\ref{eq:triplet_loss_constraint}. Formally, given $(v^\star, q^\star)$ and defining $Q$ as the set of captions in the mini-batch, the hardest negative is identified by:
\begin{equation}
    q- = argmax_{q \in Q \setminus \{q^\star\}} s(v^\star, q)\label{eq:hardneg_standard}
\end{equation}
Since $q^\star$ describes $v^\star$ in the dataset, $q^\star$ is not considered when looking for the hardest negative. Yet, not all the queries in $Q \setminus \{q^\star\}$ should be considered negative: in fact, there may be queries which correctly (or at least partially) describe $v^\star$ although the association is not present in the dataset. As an example, let $q^\star$ be `take bottle', $q_1$ `pick up bottle of wine', $q_2$ `close the fridge', and $s(v^\star, q_1) > s(v^\star, q_2)$ as in Fig.~\ref{fig:hardneg_visual}. Then, $q_1$ is the hardest negative because of two conditions: firstly, it is not $q^\star$ which, according to Eq.~\ref{eq:hardneg_standard}, makes it a possible negative; secondly, it is the closest `negative' to $v^\star$, hence selected by $argmax$. Note that $s(\cdot, \cdot)$ is tightly bound to the network state, hence after some training it might not select some of these false negatives. Nonetheless, the techniques we propose avoid these situations from the start by using the semantics of the data.

\section{Proposed method: relevance-aware online mining}

\subsection{Relevance\label{sec:relevance}}
To introduce the relevance, we start with an example. Let: ($x_1$) `pick up a flowerpot and a sunflower', ($x_2$) `pick an helianthus and a flowerpot', ($x_3$) `pot the lily in a flowerpot', ($x_4$) `put the cake in the oven'. We consider $x_2$ and $x_1$ quite similar (`helianthus' and `sunflower' are synonyms), hence $x_2$ is highly relevant; $x_3$ is slightly relevant because of `flowerpot', but the flowers and actions are different; 
and $x_4$ is irrelevant. Therefore, we want to capture semantic relations (e.g. synonyms) to determine how `similar' the two captions are, i.e. the degree of relevance. In particular, we define the relevance function $\rel(x_i, x_j)$ in terms of noun and verb classes shared among $x_i$ and $x_j$, as in Damen et al. \cite{damen2020rescaling}. Formally:
\begin{equation}\label{eq:relevance}
    \rel(x_i, x_j) = \frac{1}{2} \bigg(\frac{\vert x_i^V \cap x_j^V \vert}{\vert x_i^V \cup x_j^V \vert} + \frac{\vert x_i^N \cap x_j^N \vert}{\vert x_i^N \cup x_j^N \vert} \bigg)
\end{equation}
where $x_i^V$ and $x_i^N$ represent, respectively, the set of verb and noun classes identified in the $i$-th caption. We refer to `noun class' (or `verb class') to consider the noun (or verb) tokens which share similar semantics. When one (or both) of the inputs to $\rel$ is a video, we consider two situations. If only one caption $q_i$ is paired to the video $v_i$, we consider for $v_i$ the noun and verb classes of $q_i$. Conversely, if multiple captions are available, then we construct a word set based on the classes which appear more frequently among the different captions, as also recently done in \cite{wray2021semantic}. That is: $x_i^N = \{c^N \vert c^N \in \mathcal{D}(x_i)_{\vert \rho, N}\}$, where $c^N$ is a noun class, $\mathcal{D}(x_i)$ is the set of captions available for $x_i$, and we define $\mathcal{D}(x_i)_{\vert \rho, N}$ as the reduced set of classes for the part-of-speech $N$ which appear in at least $\rho \cdot \vert \mathcal{D}(x_i) \vert$ captions. Formally: $\mathcal{D}(x_i)_{\vert \rho, N} = \{c \, \vert \, PoS(c)=N \land \vert \{d \vert d \in  \mathcal{D}(x_i) \land c \in d\} \vert \ge \rho \cdot \vert \mathcal{D}(x_i) \vert \}$, where $PoS(\cdot)$ determines the part-of-speech of the given class. Note that $x_i^V$ is built equivalently. Finally, looking at the previous example, we can compute the following: $\rel(x_1, x_2) = 1$, $\rel(x_1, x_3) = 0.16$, and $\rel(x_1, x_4) = 0$.

\subsection{Relevance-aware online hard negative mining\label{sec:relevance_aware_hn}}
In Sec.~\ref{sec:online_hardneg_mining} we intuitively describe a limitation of current online hard negative mining. Formally, we consider $\rel$ and fix a threshold $\tau$ to determine the degree of relevance above which a caption is considered positive. Then, $\{q \, \vert \, \rel(v^\star, q) \ge \tau , q \in Q \setminus \{q^\star\}\}$ may be non empty, which may consequently lead to the selection of a caption $q-$ as negative, although it is `positive' to $v^\star$, i.e. $\rel(v^\star, q-) \ge \tau$. Considering that the triplet loss lowers the similarity of $q-$ to $v^\star$ while increasing the similarity of $q^\star$ to $v^\star$, $q-$ would be penalized although describing it correctly. To address this shortcoming, we introduce RAN, which makes the mining process aware of the relevance of the captions to the video, in order to avoid the selection of a `false negative'. We consider the following equation:
\begin{equation}
    q- = argmax_{q \in Q \setminus \{q \, \vert \, \rel(v^\star, q) \ge \tau\}} s(v^\star, q)\label{eq:hardneg_proposal}
\end{equation}
where, differently from Eq.~\ref{eq:hardneg_standard}, we employ $\{q \, \vert \, \rel(v^\star, q) \ge \tau\}$ to capture the items which should be excluded from the pool of candidate negatives.

\subsection{Relevance-aware online hard positive mining}
With the previous technique we pick high quality negative captions. Similarly, we want to select the captions which describe $v^\star$ and increase their similarity to it, to further improve the structure of the embedding space. 
To do so, we propose a two-steps approach for the relevance-aware online mining of positives. 
First of all, we compute the hardest positive $q+$ for $v^\star$, a positive caption (i.e. $\rel(v^\star, q+) \ge \tau$) which has a far too dissimilar representation when compared to $v^\star$. By following the notation used for the negative mining, this would be:
\begin{equation}
    q+ = argmin_{q \in Q \setminus \{q^\star\}} s(v^\star, q)\label{eq:hardpos_standard}
\end{equation}
but this is not optimal, as it may select as positives the easy negative captions which were not violating Eq.~\ref{eq:triplet_loss_constraint}. Hence, we propose to further employ the relevance to improve Eq.~\ref{eq:hardpos_standard}, by capturing the negative captions with $\{q \, \vert \, \rel(v^\star, q) < \tau\}$ and excluding them from the selection of the hard positives. Therefore:
\begin{equation}
    q+ = argmin_{q \in Q \setminus \{q \, \vert \, \rel(v^\star, q) < \tau\}} s(v^\star, q)\label{eq:hardpos_proposal}
\end{equation}

Then, we use these positive captions in the triplet loss, in order to increase the similarity of $v^\star$ and $q+$, while at the same time decrease the similarity with $q-$. This can be formalized as:
\begin{equation}
    L_p = max(0, \Delta_p + s(v, q-) - s(v, q+))\label{eq:triplet_loss_pos_term}
\end{equation}

Given a batch B of paired videos and captions, the final video-to-text loss is:
\begin{equation}
    \mathcal{L}_{v-t} = \frac{1}{\vert B \vert} \big(\sum_{v \in B} L_p + \sum_{v \in B} L_n \big)
\end{equation}

\section{Results}
To validate our method, we consider two large scale video-text datasets: EPIC-Kitchens-100 \cite{damen2020rescaling} and MSR-VTT \cite{xu2016msrvtt}. The former contains 67217 clips for training and 9668 for testing. Each clip is annotated with a short caption describing activities in the kitchen. Moreover, for each caption, verb and noun semantic classes are available. MSR-VTT consists of 10000 clips about multiple domains, each annotated with 20 free-form captions. We follow the official split (from \cite{xu2016msrvtt}) of 6513, 497, and 2990 clips for training, validation, and testing. To compute the semantic classes, we consider $\rho=0.25$ (see Sec.~\ref{sec:relevance}) and employ a pipeline made of spaCy, WordNet \cite{miller1995wordnet}, and Lesk algorithm \cite{lesk1986automatic} as in Wray et al. \cite{wray2021semantic}.

We consider both `text-to-video' and `video-to-text' versions of $L_n$ (Eq.~\ref{eq:triplet_loss_term}) and $L_p$ (Eq.~\ref{eq:triplet_loss_pos_term}). We employ HGR \cite{Chen_2020_CVPR} as our base model and eventually augment it with the proposed RAN and RANP. On both datasets, we perform the training for 50 epochs using a batch size of 64. For EPIC-Kitchens-100 we use TBN \cite{kazakos2019epic} features, provided alongside the dataset \cite{damen2020rescaling}. For MSR-VTT we use ImageNet-pretrained ResNet-152 features (from \cite{Chen_2020_CVPR}). 
After training, we select the best model on the validation set to perform the evaluation on the testing set.

As recently proposed by \cite{wray2021semantic}, we use Normalized Discounted Cumulative Gain (nDCG) \cite{jarvelin2002cumulated} and Mean Average Precision (mAP) \cite{baeza1999modern} for evaluation purposes. 
For MSR-VTT we only use nDCG because, due to how semantic classes are computed for its videos, the relevance values of paired captions are always lower than one, making mAP unusable. More details can be found in the Supplementary. 

\subsection{Analysis of hard negatives: relevance distribution}
\begin{figure}[t]
    \centering
    \includegraphics[width=.8\linewidth]{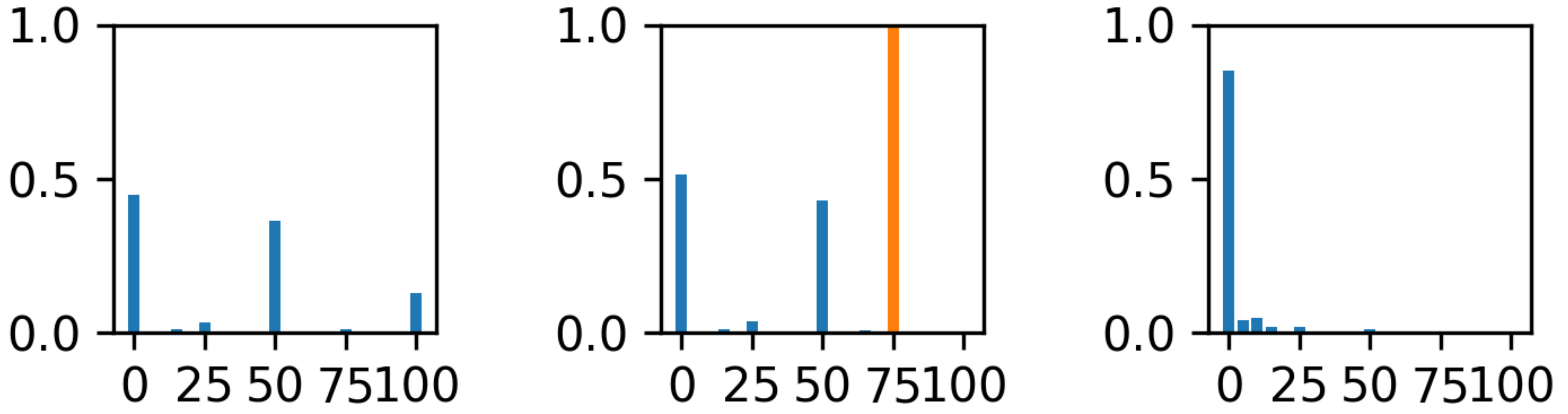}
    \caption{Distribution of relevance values of hard negatives ($\rel$ values from 0 to 100 on \textit{x} axis, relative frequency on \textit{y} axis) observed during one epoch of training (batch size 64) on EPIC-Kitchens-100 (left) and on MSR-VTT (right). In the middle, we apply the relevance-aware negative mining with $\tau=0.75$ (visualized with the orange bar).} 
    \label{fig:hardneg_distrib}
\end{figure}

In Fig.~\ref{fig:hardneg_distrib} we plot the distribution of relevance values of the hard negatives during one epoch of training. On EPIC-Kitchens-100 (Fig.~\ref{fig:hardneg_distrib} left) we observe a sizeable amount (more than 50\%) of negatives which are relevant to the query. In particular, 13\% of them have a relevance of 1. Moreover, we observe four modes for the relevance: 0 (45\%), 50 (36\%), 100 (13\%), 25 (3\%). Fig.~\ref{fig:hardneg_distrib} (middle) shows how the distribution 
changes when we apply the proposed RAN with $\tau=0.75$ (visualized with an orange bar) to improve the negatives' selection (see Sec.~\ref{sec:relevance_aware_hn}). By doing so, a lower amount of relevant items will be selected as hard negatives.

For MSR-VTT we compute a set of classes for each video based on `popular' classes which appear in the multiple descriptions (see Sec.~\ref{sec:relevance}). Consequently, the groundtruth captions for a given video may have a relevance lower than 1, making it harder to find relevant items within random mini-batches (Fig.~\ref{fig:hardneg_distrib} right). Here the top four modes are: 0 (85.2\%), 10 (4.7\%), 5 (4\%), 15 (1.7\%).

\subsection{Influence of the threshold $\tau$ on the proposed techniques}
In Table \ref{tab:ndcg_map_hp_hn} we present the results obtained by HGR when trained with the original mining technique (Eq.~\ref{eq:hardneg_standard}) followed by the usage of the proposed RAN and RANP mining strategies. We perform these experiments on both datasets, keeping $\Delta_n=\Delta_p=0.2$ (as in \cite{Chen_2020_CVPR}) and picking the values for $\tau$ from the modes observed in the previous section. In the Supplementary we show that minor changes can be observed by varying these values, although proper hyperparameter optimization is required. HGR achieves 35.9\% nDCG and 39.5\% mAP on EPIC-Kitchens-100, whereas it achieves 25.3\% nDCG on MSR-VTT. By introducing the proposed strategies, we observe consistent improvements on both datasets. 
\begin{table}[t] \centering
\begin{tabular}{|c|c|c|cc|cc|cc|}
& & & RAN & RANP & RAN & RANP & RAN & RANP \\ \hline
& $\tau$ &  & 0.75 & 0.75 & 0.40 & 0.40 & 0.15 & 0.15 \\ \hline
 \multirow{2}{*}{EPIC} & nDCG (\%) & $35.9$ & $37.4_{\uparrow 1.5}$ & $40.2_{\uparrow 4.3}$ & $48.8_{\uparrow 12.9}$ & $\mathbf{59.0_{\uparrow 23.1}}$ & $48.4_{\uparrow 12.5}$ & $58.8_{\uparrow 22.9}$ \\
 & mAP (\%) & $39.5$ & $43.1_{\uparrow 4.4}$ & $46.4_{\uparrow 6.9}$ & $46.4_{\uparrow 6.9}$ & $46.1_{\uparrow 6.6}$ & $46.5_{\uparrow 7.0}$ & $\mathbf{47.2_{\uparrow 7.7}}$ \\ \hline
 MSR-VTT & nDCG (\%) & $25.3$ & $25.2_{\uparrow 0.1}$ & - & $26.4_{\uparrow 1.1}$ & $28.0_{\uparrow 2.7}$ & $28.7_{\uparrow 3.4}$ & $\mathbf{31.1_{\uparrow 5.8}}$ \\  \hline
\end{tabular}
\caption{Performance of the baseline and the proposed RAN and RANP (improvement shown with $\,_{\uparrow X}$). We pick the values for the threshold $\tau$ close to the values of the modes (Fig.~\ref{fig:hardneg_distrib}). As $\tau$ decreases, less positives are wrongly picked as negatives. Moreover, RANP pulls closer to each video more similar captions, leading to even better performance.}
    \label{tab:ndcg_map_hp_hn}
\end{table}

First of all, lowering $\tau$ has a positive effect on both mAP and nDCG. This is likely due to the improved selection of the hard examples, which avoids several `false negatives' and leads to a stabler training. As an example, for EPIC-Kitchens-100 $\tau=0.75$ means that the examples with a relevance bigger than $\tau$ are no longer treated as possible negatives, and we observe 37.4\% nDCG (+1.5\%) and 43.1\% mAP (+4.4\%). On the other hand, for MSR-VTT the same $\tau$ leads to almost no improvements because less than 1\% of the hard negatives have a relevance higher than 0.75 (see Fig.~\ref{fig:hardneg_distrib} right). 

Secondly, by using RAN on EPIC-Kitchens-100 we observe an improvement of up to 12.9\% nDCG (48.8\%) and 7.0\% mAP (46.5\%) over the base model; on MSR-VTT the improvements measure up to 3.4\% nDCG (28.7\%). On the former we achieve great improvements thanks to the simplicity of the captions which makes it possible to easily find many relevant items and remove them from the pool of negatives. Conversely, the captions in MSR-VTT are multiple and free-form, making high relevance values rarer and thus showing lesser improvements.

Thirdly, the proposed RANP is also greatly useful. In fact, compared to only using the relevance-aware negative mining (i.e. RAN), on EPIC-Kitchens-100 we achieve further improvements, reaching 59.0\% nDCG with $\tau=0.40$ (+23.1\% over the baseline) and 47.2\% mAP with $\tau=0.15$ (+7.7\%). On MSR-VTT we also observe considerable improvements in terms of nDCG (up to +5.8\%). With RANP, during training we ensure that its original caption is pulled near the video, but also other captions which describe its visual contents, further improving the quality of the ranked lists. Based on these observations, in the following experiments we use $\tau=0.15$ for EPIC-Kitchens-100 and $\tau=0.10$ for MSR-VTT.

\subsection{Comparison with state-of-the-art}
In Tables \ref{tab:sota_ndcg_map_ek100} and \ref{tab:sota_ndcg_map_msr} we report the results we obtain with HGR \cite{Chen_2020_CVPR} augmented by the proposed techniques on both EPIC-Kitchens-100 and MSR-VTT, and propose a comparison to other popular methods.

\begin{table}[t]
    \centering
    \begin{tabular}{|l|ccc|ccc|}
        & \multicolumn{6}{c|}{EPIC-Kitchens-100} \\ \hline
        & \multicolumn{3}{c|}{nDCG (\%)} & \multicolumn{3}{c|}{mAP (\%)} \\ \hline
        Model & t2v & v2t & avg & t2v & v2t & avg \\ \hline
        HGR \cite{Chen_2020_CVPR} & 37.9 & 41.2 & 35.9 & 35.7 & 36.1 & 39.5 \\ 
        MME \cite{wray2019fine} & 46.9 & 50.0 & 48.5 & 34.0 & 43.0 & 38.5 \\ 
        JPoSE \cite{wray2019fine} & 51.5 & 55.5 & 53.5 & 38.1 & 49.9 & 44.0 \\ 
        Hao et al. \cite{Damen2021CHALLENGES} & 51.8 & 55.3 & 53.5 & 38.5 & 50.0 & 44.2 \\ \hline
        RAN & 47.1 & 49.7 & 48.4 & 43.1 & 49.9 & 46.5 \\
        \textbf{RANP} & $\mathbf{56.5_{\uparrow 4.7}}$ & $\mathbf{61.2_{\uparrow 5.7}}$ & $\mathbf{58.8_{\uparrow 5.3}}$ & $\mathbf{42.3_{\uparrow 3.8}}$ & $\mathbf{52.0_{\uparrow 2.0}}$ & $\mathbf{47.2_{\uparrow 3.0}}$ \\
        \hline
    \end{tabular}
    \caption{Comparison with the baseline (HGR \cite{Chen_2020_CVPR}) and state-of-the-art methods for EPIC-Kitchens-100 (results for MME and JPoSE are from \cite{damen2020rescaling}). 
    By using RAN we achieve competitive mAP. With RANP, which identifies both negatives and positives via relevance, we achieve state-of-the-art results on mAP and nDCG simultaneously.} 
    \label{tab:sota_ndcg_map_ek100}
\end{table}
\textbf{EPIC-Kitchens-100.} In Table \ref{tab:sota_ndcg_map_ek100} we compare to MME and JPoSE, proposed by Wray et al. \cite{wray2019fine} and used in \cite{damen2020rescaling} as the baselines for the challenge. We include Hao et al. \cite{Damen2021CHALLENGES} which is the current the state-of-the-art (53.5\% nDCG and 44.2\% mAP). We observe considerable improvements (+5.3\% nDCG and +3.0\% mAP) by using the proposed RANP, leading us to a new state-of-the-art result (58.8\% nDCG and 47.2\% mAP). Moreover, such an improvement is observed when looking both at the average and at task-level values (text-to-video and video-to-text), giving a clear evidence of the usefulness of the proposed techniques.

\textbf{MSR-VTT.} For MSR-VTT, we compare to MoEE (Miech et al. \cite{miech2018learning}) and CE (Liu et al. \cite{liu2019use}). To have a fair comparison, in Table \ref{tab:sota_ndcg_map_msr} we evaluate them using only appearance features within the open source codebase of \cite{liu2019use}. Then, we include the results for HGR and for the proposed techniques. 
CE and MoEE present higher nDCG rates (respectively 29.4\% and 29.0\%) than the base HGR (25.3\%), showing that CE and MoEE compute higher quality ranked lists than our baseline. If we augment HGR with the proposed relevance-aware hard negative mining (RAN) we observe an improvement of +3.4\% nDCG (28.7\%). This is due to having less positive captions wrongly selected as negatives during training. If we also introduce the proposed variant for hard positives (using the full RANP), we observe a further improvement, leading to an overall margin of +2.2\% over CE (31.6\% versus 29.4\%). With the introduction of this second technique, less irrelevant captions are retrieved at the top of the ranked list. 
\begin{table}[t]
    \centering
    \begin{tabular}{|c|c|c|cc|c|cc|}
 &  & Model & CE & MoEE & HGR & RAN & \textbf{RANP} \\ \hline
\multirow{3}{*}{MSR-VTT} & \multirow{3}{*}{nDCG (\%)} & t2v & 28.9 & 28.4 & 24.6 & 27.4 & $\mathbf{29.1_{\uparrow 0.2}}$ \\
 &  & v2t & 30.0 & 29.5 & 26.1 & 30.1 & $\mathbf{34.1_{\uparrow 4.1}}$ \\
 &  & avg & 29.4 & 29.0 & 25.3 & 28.7 & $\mathbf{31.6_{\uparrow 2.2}}$ \\ \hline
   \end{tabular}
    \caption{For a fair comparison with MoEE \cite{miech2018learning} and CE \cite{liu2019use} we evaluate their performance while using appearance features and the code provided by the authors. Then we evaluate the baseline (HGR) and the proposed techniques.} 
    \label{tab:sota_ndcg_map_msr}
\end{table}


\section{Conclusions}
Video retrieval methods are usually trained using a contrastive loss, such as the triplet loss \cite{schroff2015facenet}. During training, the negatives are selected among the captions or videos which are not associated in the dataset. In this paper, we showed that the typical formulation used to mine the negatives also selects captions which partially describe the input video. To emphasize the importance of this selection step, we proposed the relevance-aware online hard negative mining, which uses a relevance function to separate positive and negative items. Furthermore, in the video retrieval community positive examples hardly are mined because proper labels are usually absent. To this end, we also proposed the relevance-aware online hard positive mining. Finally, we gave empirical evidence of the strength of the proposed techniques by applying them on a deep learning model (HGR \cite{Chen_2020_CVPR}) and testing it on two benchmark datasets: the recently released EPIC-Kitchens-100 \cite{damen2020rescaling} and MSR-VTT \cite{xu2016msrvtt}. In both cases, the application of the proposed techniques leads to considerable improvements and state-of-the-are results.

\section{Appendix}
\subsection{Evaluation metrics\label{sup:metrics}}
As recently proposed by \cite{wray2021semantic}, we use the Normalized Discounted Cumulative Gain (nDCG) \cite{jarvelin2002cumulated} and the Mean Average Precision (mAP) \cite{baeza1999modern} for evaluation purposes. As is done in \cite{jarvelin2002cumulated,damen2020rescaling}, we define the DCG in terms of the relevance (see Sec. 4.1): given an item $a$ and a ranked list of queries $Q$ (restricted to the top $N_r$ relevant items), the DCG is given by
\begin{equation}
    DCG(a, Q) = \sum_{k=1}^{N_r} \frac{\rel(a, q_k)}{log_2 (k+1)}
\end{equation}
whereas the nDCG is obtained by normalizing the DCG with respect to IDCG, which is the optimal value obtained by sorting the ranked list following a descending order of relevance values:
\begin{equation}
    nDCG(a, Q) = \frac{DCG(a, Q)}{IDCG(a, Q)}
\end{equation}
Let $N$ be the length of the ranking list (consisting of both the irrelevant items and the $N_r$ relevant items), $P(k)$ is the Precision at k \cite{baeza1999modern}. Let $r(k)=1$ when $\rel(a, q_k)=1$, and $r(k)=0$ otherwise (that is, $r(\cdot)$ is an indicator function of binary relevance). Then, for an item $a$, the Average Precision (AP) is:
\begin{equation}
    AP(a) = \frac{\sum_{k=1}^N P(k) \cdot r(k)}{N_r}    
\end{equation}
whereas the mAP is given by the mean over all the items. An important difference between the two metrics lies in how they consider the relevance: while it is a continuous function in [0, 1] for the DCG, the AP only considers a binary definition of the relevance.

\subsection{Different margins for negatives and positives\label{sup:margins}}
In Sec. 5.2 of the main paper, we used the same value for both margins of the contrastive loss terms ($\Delta_n$ in Eq. 1 and $\Delta_p$ in Eq. 8). Yet, changing their value may lead to embedding spaces which are organized differently which, in turn, may affect the final performance. To explore this inquiry, we fix $\Delta_n=0.2$ and vary $\Delta_p$ in $\{0.10, 0.15, 0.20, 0.25, 0.30\}$. The results are shown in Table \ref{tab:diff_margins}. Note that for $\tau$ we use the values leading to the best average nDCG in Sec. 5.2 of the main paper: for EPIC-Kitchens-100 we use $\tau=0.4$, whereas for MSR-VTT we use $\tau=0.10$. As can be seen, changing $\Delta_p$ has limited impact on the nDCG performance on EPIC-Kitchens-100, leading to up to -0.3\% when $\Delta_p$ is decreased and up to +0.2\% when it is increased. On the other hand, by decreasing $\Delta_p$ an improvement of up to +1.0\% is observed for the mAP. Conversely, on MSR-VTT we observe lesser improvements in terms of nDCG (up to +0.5\%).

\begin{table}[]
    \centering
    \begin{tabular}{|cc|cc|c|}
        & & \multicolumn{2}{c|}{EPIC} & \multicolumn{1}{c|}{MSR-VTT} \\ \hline
        $\Delta_n$ & $\Delta_p$ & nDCG (\%) & mAP (\%) & nDCG (\%) \\ \hline
        0.20 & 0.10 & 58.7 & \textbf{47.1} & 31.2 \\
        0.20 & 0.15 & 59.0 & 46.6 & 31.3  \\ \hline
        0.20 & 0.20 & 59.0 & 46.1 & 31.1 \\ \hline
        0.20 & 0.25 & \textbf{59.2} & 46.0 & \textbf{31.6} \\
        0.20 & 0.30 & 59.0 & 45.6 & \textbf{31.6} \\ \hline
    \end{tabular}
    \caption{We keep $\Delta_n=0.20$ and vary $\Delta_p$ in \{0.10, 0.15, 0.20, 0.25, 0.30\}. On EPIC-Kitchens-100 we observe up to +1\% mAP when compared to the default values for the margins (i.e. 0.20), whereas on MSR-VTT we observe smaller variations.}
    \label{tab:diff_margins}
\end{table}

\noindent\textbf{Acknowledgements.}
We gratefully acknowledge the support from Amazon AWS Machine Learning Research Awards (MLRA) and NVIDIA AI Technology Centre (NVAITC), EMEA. We acknowledge the CINECA award under the ISCRA initiative, which provided  computing resources for this work.

\bibliographystyle{splncs04}
\bibliography{biblio}

\begin{thebibliography}{10}
\providecommand{\url}[1]{\texttt{#1}}
\providecommand{\urlprefix}{URL }
\providecommand{\doi}[1]{https://doi.org/#1}

\bibitem{arandjelovic2016netvlad}
Arandjelovic, R., Gronat, P., Torii, A., Pajdla, T., Sivic, J.: Netvlad: Cnn
  architecture for weakly supervised place recognition. In: Proceedings of the
  IEEE CVPR. pp. 5297--5307 (2016)

\bibitem{baeza1999modern}
Baeza-Yates, R., Ribeiro-Neto, B., et~al.: Modern information retrieval,
  vol.~463. ACM press New York (1999)

\bibitem{Chen_2020_CVPR}
Chen, S., Zhao, Y., Jin, Q., Wu, Q.: Fine-grained video-text retrieval with
  hierarchical graph reasoning. In: Proceedings of the IEEE/CVF CVPR (June
  2020)

\bibitem{chen2020simple}
Chen, T., Kornblith, S., Norouzi, M., Hinton, G.: A simple framework for
  contrastive learning of visual representations. In: International conference
  on machine learning. pp. 1597--1607. PMLR (2020)

\bibitem{chen2017beyond}
Chen, W., Chen, X., Zhang, J., Huang, K.: Beyond triplet loss: a deep
  quadruplet network for person re-identification. In: Proceedings of the IEEE
  CVPR. pp. 403--412 (2017)

\bibitem{croitoru2021teachtext}
Croitoru, I., Bogolin, S.V., Leordeanu, M., Jin, H., Zisserman, A., Albanie,
  S., Liu, Y.: Teachtext: Crossmodal generalized distillation for text-video
  retrieval. In: Proceedings of the IEEE/CVF ICCV. pp. 11583--11593 (2021)

\bibitem{damen2020rescaling}
Damen, D., Doughty, H., Farinella, G.M., Furnari, A., Kazakos, E., Ma, J.,
  Moltisanti, D., Munro, J., Perrett, T., Price, W., et~al.: Rescaling
  egocentric vision. IJCV  (2021)

\bibitem{Damen2021CHALLENGES}
Damen, D., Fragomeni, A., Munro, J., Perrett, T., Whettam, D., Wray, M.,
  Furnari, A., Farinella, G.M., Moltisanti, D.: Epic-kitchens-100- 2021
  challenges report. Tech. rep., University of Bristol (2021)

\bibitem{dong2021dual}
Dong, J., Li, X., Xu, C., Yang, X., Yang, G., Wang, X., Wang, M.: Dual encoding
  for video retrieval by text. IEEE Transactions on Pattern Analysis and
  Machine Intelligence  (2021)

\bibitem{dzabraev2021mdmmt}
Dzabraev, M., Kalashnikov, M., Komkov, S., Petiushko, A.: Mdmmt: Multidomain
  multimodal transformer for video retrieval. In: Proceedings of the IEEE/CVF
  CVPR. pp. 3354--3363 (2021)

\bibitem{gabeur2020multi}
Gabeur, V., Sun, C., Alahari, K., Schmid, C.: Multi-modal transformer for video
  retrieval. In: Proceedings of the IEEE ECCV. Springer (2020)

\bibitem{gordo2017beyond}
Gordo, A., Larlus, D.: Beyond instance-level image retrieval: Leveraging
  captions to learn a global visual representation for semantic retrieval. In:
  Proceedings of the IEEE CVPR. pp. 6589--6598 (2017)

\bibitem{gutmann2010noise}
Gutmann, M., Hyv{\"a}rinen, A.: Noise-contrastive estimation: A new estimation
  principle for unnormalized statistical models. In: Proceedings of the
  thirteenth international conference on artificial intelligence and
  statistics. pp. 297--304. JMLR Workshop and Conference Proceedings (2010)

\bibitem{hadsell2006dimensionality}
Hadsell, R., Chopra, S., LeCun, Y.: Dimensionality reduction by learning an
  invariant mapping. In: 2006 IEEE Computer Society CVPR (CVPR'06). vol.~2, pp.
  1735--1742. IEEE (2006)

\bibitem{harwood2017smart}
Harwood, B., Kumar~BG, V., Carneiro, G., Reid, I., Drummond, T.: Smart mining
  for deep metric learning. In: Proceedings of the IEEE ICCV. pp. 2821--2829
  (2017)

\bibitem{he2020momentum}
He, K., Fan, H., Wu, Y., Xie, S., Girshick, R.: Momentum contrast for
  unsupervised visual representation learning. In: Proceedings of the IEEE/CVF
  CVPR. pp. 9729--9738 (2020)

\bibitem{hermans2017defense}
Hermans, A., Beyer, L., Leibe, B.: In defense of the triplet loss for person
  re-identification. arXiv preprint arXiv:1703.07737  (2017)

\bibitem{jarvelin2002cumulated}
J{\"a}rvelin, K., Kek{\"a}l{\"a}inen, J.: Cumulated gain-based evaluation of ir
  techniques. ACM Transactions on Information Systems (TOIS)  \textbf{20}(4),
  422--446 (2002)

\bibitem{jiang2019svd}
Jiang, Q.Y., He, Y., Li, G., Lin, J., Li, L., Li, W.J.: Svd: A large-scale
  short video dataset for near-duplicate video retrieval. In: Proceedings of
  the IEEE/CVF ICCV. pp. 5281--5289 (2019)

\bibitem{kazakos2019epic}
Kazakos, E., Nagrani, A., Zisserman, A., Damen, D.: Epic-fusion: Audio-visual
  temporal binding for egocentric action recognition. In: Proceedings of the
  IEEE/CVF ICCV. pp. 5492--5501 (2019)

\bibitem{lesk1986automatic}
Lesk, M.: Automatic sense disambiguation using machine readable dictionaries:
  how to tell a pine cone from an ice cream cone. In: Proceedings of the 5th
  annual international conference on Systems documentation. pp. 24--26 (1986)

\bibitem{liu2019use}
Liu, Y., Albanie, S., Nagrani, A., Zisserman, A.: Use what you have: Video
  retrieval using representations from collaborative experts. BMVC  (2019)

\bibitem{miech2018learning}
Miech, A., Laptev, I., Sivic, J.: Learning a text-video embedding from
  incomplete and heterogeneous data. arXiv preprint arXiv:1804.02516  (2018)

\bibitem{miller1995wordnet}
Miller, G.A.: Wordnet: a lexical database for english. Communications of the
  ACM  \textbf{38}(11),  39--41 (1995)

\bibitem{mithun2018learning}
Mithun, N.C., Li, J., Metze, F., Roy-Chowdhury, A.K.: Learning joint embedding
  with multimodal cues for cross-modal video-text retrieval. In: Proceedings of
  the 2018 ACM on International Conference on Multimedia Retrieval. pp. 19--27
  (2018)

\bibitem{pan2021videomoco}
Pan, T., Song, Y., Yang, T., Jiang, W., Liu, W.: Videomoco: Contrastive video
  representation learning with temporally adversarial examples. In: Proceedings
  of the IEEE/CVF CVPR. pp. 11205--11214 (2021)

\bibitem{qian2021spatiotemporal}
Qian, R., Meng, T., Gong, B., Yang, M.H., Wang, H., Belongie, S., Cui, Y.:
  Spatiotemporal contrastive video representation learning. In: Proceedings of
  the IEEE/CVF CVPR. pp. 6964--6974 (2021)

\bibitem{schroff2015facenet}
Schroff, F., Kalenichenko, D., Philbin, J.: Facenet: A unified embedding for
  face recognition and clustering. In: Proceedings of the IEEE CVPR. pp.
  815--823 (2015)

\bibitem{sohn2016improved}
Sohn, K.: Improved deep metric learning with multi-class n-pair loss objective.
  In: Advances in neural information processing systems. pp. 1857--1865 (2016)

\bibitem{suh2019stochastic}
Suh, Y., Han, B., Kim, W., Lee, K.M.: Stochastic class-based hard example
  mining for deep metric learning. In: Proceedings of the IEEE/CVF CVPR. pp.
  7251--7259 (2019)

\bibitem{wang2021t2vlad}
Wang, X., Zhu, L., Yang, Y.: T2vlad: global-local sequence alignment for
  text-video retrieval. In: Proceedings of the IEEE/CVF CVPR. pp. 5079--5088
  (2021)

\bibitem{wray2021semantic}
Wray, M., Doughty, H., Damen, D.: On semantic similarity in video retrieval.
  In: Proceedings of the IEEE/CVF CVPR. pp. 3650--3660 (2021)

\bibitem{wray2019fine}
Wray, M., Larlus, D., Csurka, G., Damen, D.: Fine-grained action retrieval
  through multiple parts-of-speech embeddings. In: Proceedings of the IEEE
  ICCV. pp. 450--459 (2019)

\bibitem{xu2016msrvtt}
Xu, J., Mei, T., Yao, T., Rui, Y.: Msr-vtt: A large video description dataset
  for bridging video and language. In: Proceedings of the IEEE CVPR. pp.
  5288--5296 (2016)

\bibitem{xuan2020hard}
Xuan, H., Stylianou, A., Liu, X., Pless, R.: Hard negative examples are hard,
  but useful. In: Proceedings of the IEEE ECCV. pp. 126--142 (2020)

\bibitem{xuan2020improved}
Xuan, H., Stylianou, A., Pless, R.: Improved embeddings with easy positive
  triplet mining. In: Proceedings of the IEEE/CVF Winter Conference on
  Applications of Computer Vision. pp. 2474--2482 (2020)

\end{thebibliography}

\end{document}